\documentclass[runningheads]{llncs}
    
    \usepackage{eccv}
    
    \usepackage{eccvabbrv}
    
    \usepackage{graphicx}
    \usepackage{booktabs}
    \usepackage{amssymb} 
    \usepackage{multirow}
    \usepackage{float}
    \usepackage[accsupp]{axessibility}  

    \usepackage{hyperref}
    
    \usepackage{orcidlink}

\begin{document}
    
    \title{Inst4DGS: Instance-Decomposed 4D Gaussian Splatting with Multi-Video Label Permutation Learning} 
    
    \titlerunning{Inst4DGS}
    



    \author{
    Yonghan Lee \and
    Dinesh Manocha
    }
    
    \authorrunning{Y. Lee and D. Manocha}
    
    \institute{
    University of Maryland, College Park, USA \\
    \email{\{lyhan12,dmanocha\}@umd.edu}
    }
    \maketitle
    
    \begin{center}
    \includegraphics[width=\textwidth]{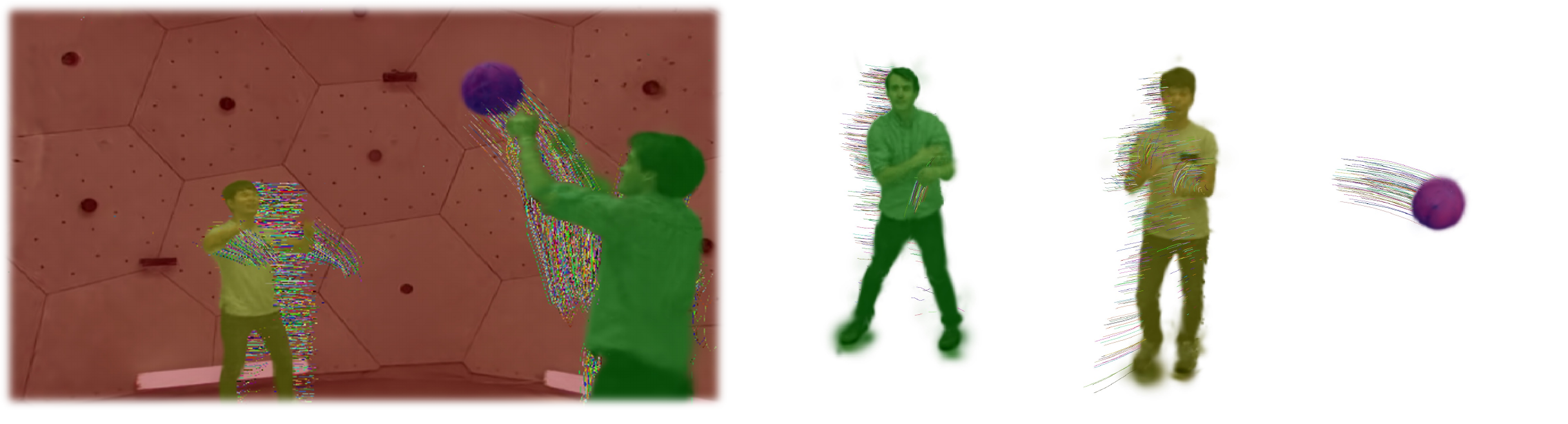}
    \captionof{figure}{
    We introduce a novel instance-decomposed 4D Gaussian Splatting (4DGS) approach for long-term, identity-consistent tracking. Given multi-view videos, our method reconstructs a 4D scene representation (left) and decomposes it into per-instance 4DGS tracks (right). By integrating instance segmentation with 4DGS optimization through per-instance motion bases, we enable efficient motion modeling and stable long-horizon tracking for each scene object.
    }
    \label{fig:intro}
    \vspace{-1.3em}
    \end{center}

    \begin{abstract}
    We present Inst4DGS, an instance-decomposed 4D Gaussian Splatting (4DGS) approach with long-horizon per-Gaussian trajectories. While dynamic 4DGS has advanced rapidly, instance-decomposed 4DGS remains underexplored, largely due to the difficulty of associating inconsistent instance labels across independently segmented multi-view videos. We address this challenge by introducing per-video label-permutation latents that learn cross-video instance matches through a differentiable Sinkhorn layer, enabling direct multi-view supervision with consistent identity preservation. This explicit label alignment yields sharp decision boundaries and temporally stable identities without identity drift. To further improve efficiency, we propose instance-decomposed motion scaffolds that provide low-dimensional motion bases per object for long-horizon trajectory optimization. Experiments on Panoptic Studio and Neural3DV show that Inst4DGS jointly supports tracking and instance decomposition while achieving state-of-the-art rendering and segmentation quality. On the Panoptic Studio dataset, Inst4DGS improves PSNR from 26.10 to 28.36, and instance mIoU from 0.6310 to 0.9129, over the strongest baseline.

    \keywords{4D Gaussian Splatting (4DGS) \and 4D instance segmentation \and scene tracking \and neural scene rendering}
    \end{abstract}

    \section{Introduction}
    \label{sec:intro}
    Real-world scenes are inherently dynamic, with numerous moving objects and people. Understanding and interacting with such environments requires reconstructing not only scene geometry and appearance, but also the motion and identity of individual objects over time. Instance-aware 4D scene reconstruction and tracking can address this need by decomposing dynamic scenes into objects and trajectories~\cite{ji2024segment4dgaussians,li2026trasetrackingfree4dsegmentation,split4d2025,ye2024gaussiangrouping},  which is crucial for downstream applications in VR/XR, robotics, and autonomous driving.

    

    \begin{figure*}[!t]
    \centering
    \includegraphics[width=0.9\linewidth]{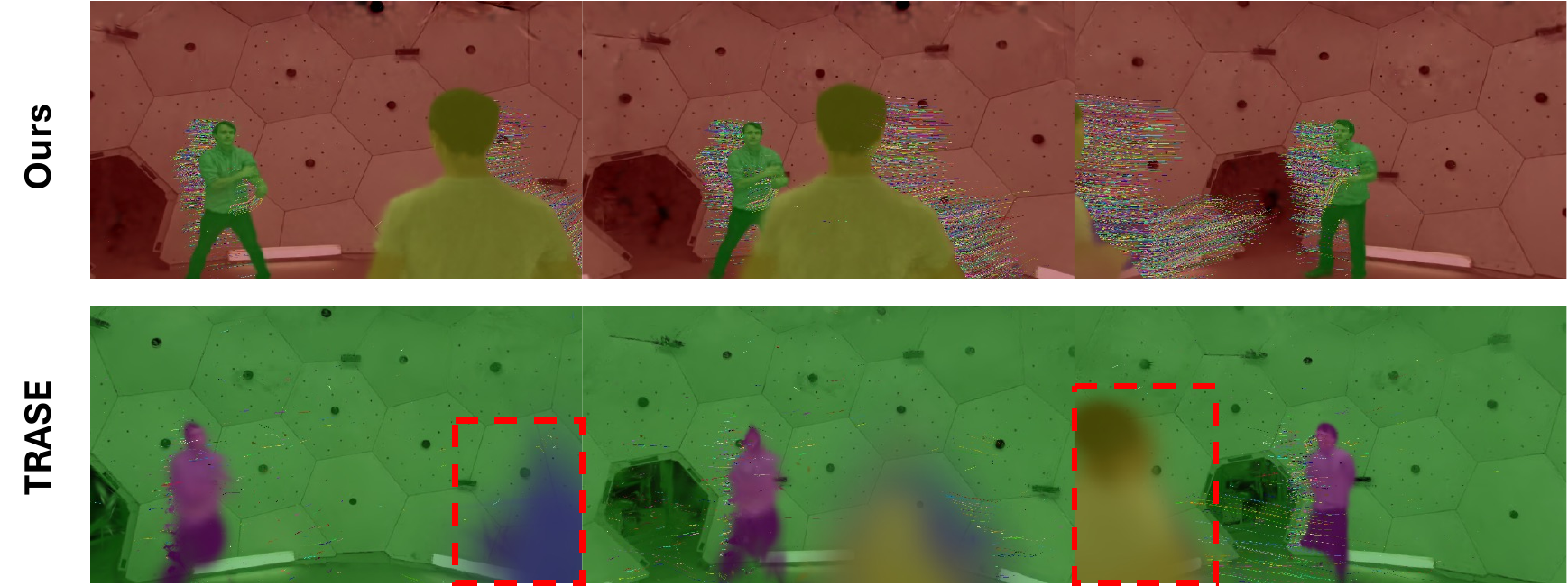}
    \caption{Temporally consistent identity reconstruction. Our method maintains stable instance identities over time while achieving high-fidelity tracking and rendering. In contrast, TRASE \cite{li2026trasetrackingfree4dsegmentation} exhibits limited motion tracking and identity drift (red box).}
    \label{fig:identity_shift}
\end{figure*}


    Recently, 3D Gaussian Splatting (3DGS) has emerged as a dominant paradigm for high-fidelity scene reconstruction, demonstrating efficient differentiable rendering with explicit 3D Gaussian primitives~\cite{kerbl20233dgs}. An important advantage of such explicit primitives is their natural extension to dynamic scene modeling, where motion can be represented by associating trajectories with individual Gaussians. Building on this idea, various 4D Gaussian Splatting (4DGS) \cite{wu20244dgs, Wu2024CVPR4dgs, duisterhof2023deformgs, li2026trasetrackingfree4dsegmentation, luiten2023dynamic3dgs} approaches have been proposed with different motion parameterizations. For example, several methods \cite{yang2023deformable3dgs, li2026trasetrackingfree4dsegmentation, labe2024dgd} employ deformation fields \cite{yang2023deformable3dgs} to model the motion of static Gaussians across time. Other approaches represent dynamic scenes with temporally defined 4D Gaussian primitives~\cite{li2024stgs, yang2023real}. However, they primarily target rendering quality and lack explicit trajectory models for long-term tracking.
    
    
    Another line of work explicitly models long-term trajectories of Gaussians for 4D reconstruction. For instance, Dynamic3DGS~\cite{luiten2023dynamic3dgs} jointly performs long-term tracking and reconstruction by sequentially optimizing the trajectories of persistent Gaussians. However, optimizing trajectories for all Gaussians independently requires extensive computation and lacks a low-dimensional or semantic structure for modeling coherent object motion. Moreover, existing 4DGS tracking approaches remain object-agnostic, and instance-aware 4DGS representations with long-term persistent tracking are still largely unexplored.
    
    Research on instance-decomposed 4D scene reconstruction is relatively limited, with only a few recent works~\cite{ji2024segment4dgaussians, li2026trasetrackingfree4dsegmentation, split4d2025}. A central challenge in 4D scene segmentation is the difficulty of leveraging inconsistent instance labels produced by video segmentation models \cite{carion2025sam3, cheng2023deva} across multiple views \cite{split4d2025, li2026trasetrackingfree4dsegmentation}. Since each video is typically processed independently, the resulting segmentation maps have different label spaces across cameras. To address this issue, prior works such as~\cite{li2026trasetrackingfree4dsegmentation, split4d2025} adopt contrastive learning to avoid direct supervision from inconsistent labels, while SA4D \cite{ji2024segment4dgaussians} relies on only single-view supervision.
    
    However, contrastive feature learning produces implicit clustering boundaries, which do not guarantee consistent object identities across space and time, as also discussed in~\cite{ji2024segment4dgaussians}. As illustrated in Fig.~\ref{fig:identity_shift}, the implicitly learned feature field in TRASE \cite{li2026trasetrackingfree4dsegmentation} can exhibit identity drift, since instance assignments are not directly supervised by the provided segmentation labels. Furthermore, these methods do not explicitly model long-term object trajectories, and therefore cannot produce per-instance persistent 4DGS tracks.
    

    
    \noindent {\bf Main Results:}
    We present a novel, general solution for \textit{instance-decomposed}, \textit{long-term persistent} 4D scene tracking and reconstruction, while studying the synergy between object instance decomposition and motion modeling.  Our approach is based on two key ideas:  (1) directly leverage per-video segmentation maps to enforce identity preservation of long-term persistent Gaussians for each object; (2) instance grouping provides useful structure for motion modeling, enabling more efficient optimization of object motion in 4D Gaussian representations.
    We propose Inst4DGS, an instance-decomposed 4D Gaussian Splatting framework that supports long-term persistent trajectory tracking with consistent object identities. To enable direct supervision from per-view video segmentation maps, we introduce \textit{per-video latent permutation} variables that model label correspondences across different videos and can be optimized in a differentiable manner. We employ a differentiable Sinkhorn layer \cite{mena2018gumbelsinkhorn} to enforce one-to-one matching constraints while allowing gradient-based learning of permutation variables. These permutation latents form a compact set of learnable per-video parameters that efficiently resolve cross-view label inconsistencies, enabling direct supervision from multiple segmentation maps and ensuring identity preservation for each object instance. Inspired by \cite{lei2025mosca}, to efficiently model object motion, we introduce an \textit{instance-decomposed motion scaffold} that serves as motion bases for groups of Gaussians belonging to the same object. This representation allows efficient optimization of trajectories while maintaining coherent object motion. Our main contributions are summarized as follows:
    
    \begin{itemize}
    
    \item We propose \textit{Inst4DGS}, an instance-decomposed 4D Gaussian Splatting framework that jointly enables long-term persistent tracking and instance-level scene reconstruction. This unique capability of our pipeline is highlighted in Table \ref{tab:photometric_multidataset_pan_neural} and Fig. \ref{fig:intro}.
    
    \item We introduce a \textit{per-video permutation latent representation} with Sinkhorn normalization to resolve cross-video label inconsistencies, enabling direct supervision from multi-view segmentation maps while preserving consistent object identities. In contrast to concurrent baseline TRASE, our method does not suffer from the temporal identity shift (Fig. \ref{fig:identity_shift}).
    
    \item We propose an \textit{instance-decomposed motion scaffold} that provides a compact motion structure for groups of Gaussians, significantly improving the efficiency of long-horizon trajectory optimization.
    
    \item Extensive experiments on \textit{Panoptic Studio} and \textit{Neural3DV} demonstrate strong performance in photometric rendering (PSNR: 28.60), dynamic segmentation (mIoU: 0.9569), and instance segmentation (mIoU: 0.9129), outperforming current methods. On the Panoptic Studio dataset, Inst4DGS improves PSNR from $26.10$ to $28.36$, and instance mIoU from $0.6310$ to $0.9129$, over the strongest baselines. More comparisons are given in Tables 2 and 3.
    \end{itemize}

    
    
    
    \section{Related Work}
    \subsection{4D Gaussian Splatting}
    3D Gaussian Splatting (3DGS)~\cite{kerbl20233dgs} is a dominant explicit representation for high-fidelity view synthesis due to differentiable rasterization and rendering. Dynamic extensions to 3DGS (4DGS) largely fall into three families: canonical-deformation methods (e.g., Deformable 3D Gaussians~\cite{yang2023deformable3dgs}), which provide strong photometric quality but can struggle with long-term complex motion~\cite{split4d2025}; spatio-temporal 4D primitives (e.g., 4D Gaussian Splatting~\cite{wu20244dgs}, STGS~\cite{li2024stgs}), which prioritize rendering efficiency but have limited temporal support for explicit long-horizon trajectories; and explicit persistent-trajectory models (e.g., Dynamic 3D Gaussians~\cite{luiten2023dynamic3dgs}), which improve temporal coherence at a high optimization cost. Recent motion-basis methods improve the efficiency of explicit-trajectory 4DGS~\cite{som2024, lei2025mosca}, but are mainly designed for monocular or single-video settings, making extension to multi-video optimization non-trivial. Inst4DGS addresses this gap by leveraging 4D instance segmentation to initialize per-instance motion bases for efficient multi-video explicit-trajectory reconstruction.

    \subsection{3D Semantic Scene Reconstruction}
    A parallel line of work lifts 2D semantic priors into 3D neural scene representations, starting from NeRF~\cite{mildenhall2021nerf}. Representative feature-lifting methods include Neural Feature Fusion Fields~\cite{tschernezki2022neuralfeaturefusionfields}, LERF~\cite{lerf2023}, Distilled Feature Fields~\cite{shen2023distilledfeaturefields}, Feature-3DGS~\cite{zhou2024feature}, Gaussian Grouping~\cite{ye2024gaussiangrouping}, and FMGS~\cite{zuo2024fmgs}. Panoptic Lifting~\cite{siddiqui2023panopticlifting} is especially relevant because it explicitly handles cross-view instance-ID assignment. However, these methods mostly target static or quasi-static 3D scenes and do not address long-horizon 4D identity consistency under multi-video label-space mismatch. In our approach, we leverage direct label supervision to learn identity fields for dynamic 4D Gaussians, similar to Gaussian Grouping~\cite{ye2024gaussiangrouping}, while extending the formulation to multi-video 4D settings with implicit instance label matching between multiple video views.

    \subsection{Instance-Decomposed 4D Gaussian Splatting}
    Recent works begin to address instance-aware dynamic 4D reconstruction, including Segment Any 4D Gaussians (SA4D) \cite{ji2024segment4dgaussians}, TRASE \cite{li2026trasetrackingfree4dsegmentation}, and Split4D \cite{split4d2025}. SA4D \cite{ji2024segment4dgaussians} adopts a GaussianGrouping-style \cite{ye2024gaussiangrouping} instance field representation to learn instance embeddings from monocular video segmentation maps. TRASE \cite{li2026trasetrackingfree4dsegmentation} and Split4D \cite{split4d2025} learn instance-aware 4DGS from multiple per-video segmentation maps using contrastive objectives. Although contrastive learning avoids direct dependence on inconsistent label IDs by learning pairwise similarity, it induces implicit clustering boundaries and can still cause identity drift, as also discussed in SA4D \cite{ji2024segment4dgaussians}. Instead, we explicitly model cross-view label assignment and directly supervise canonical labels from segmentation maps. Inspired by Sinkhorn-based permutation learning \cite{mena2018gumbelsinkhorn}, we introduce learnable permutation latents to enforce consistent identities across views and time.

    \begin{figure*}[!t]
        \centering
        \includegraphics[width=1.0\linewidth]{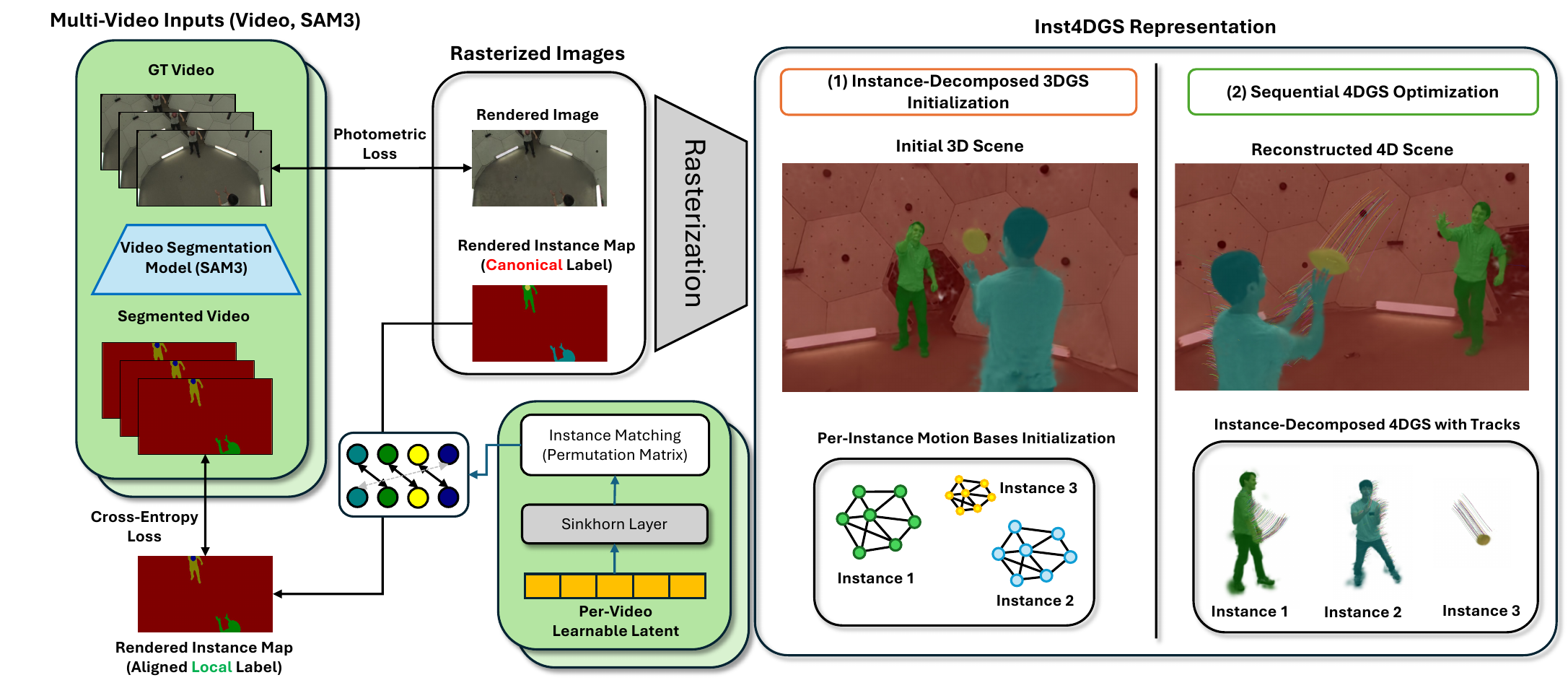}
        \caption{\textbf{Inst4DGS Pipeline:} Given multi-view videos and per-video segmentation labels, our method reconstructs an instance-decomposed 4DGS scene with long-term per-instance trajectories. The pipeline has two stages: (1) instance-decomposed 3DGS initialization and (2) sequential 4D optimization. A per-video learnable latent differentiably aligns inconsistent instance labels across views, and per-instance motion bases provide a low-dimensional motion scaffold for efficient optimization.}
        \label{fig:method}
    \end{figure*}

    \section{Inst4DGS Pipeline}
    \subsubsection{Pipeline Summary}
    Our pipeline consists of initialization step and optimization step, along with our novel components for per-video permutation learning and instance-decomposed motion bases.
    We illustrate the Inst4DGS pipeline in Fig.~\ref{fig:method}. Given multi-view RGB videos and their per-video segmentation masks, our method reconstructs an instance-decomposed 4DGS representation with long-term trajectories. The pipeline has two stages. First, we optimize an instance-decomposed 3DGS at the initial time step (Sec.~4) while implicitly matching labels across videos. We then initialize per-instance motion bases and optimize their trajectories via sequential 4DGS optimization (Sec.~5).
    
    \subsubsection{4DGS Representation}
    Inspired by GaussianGrouping \cite{ye2024gaussiangrouping} and Dynamic3DGS \cite{luiten2023dynamic3dgs}, we adopt a 4D Gaussian Splatting (4DGS) representation where each Gaussian follows a per-timestep trajectory $\mu_i(t)$ while retaining standard 3DGS attributes.
    Each Gaussian $G_i$ is parameterized as
    \[
    G_i = \{c_i, o_i, s_i, r_i, \mu_i(t), f_i\},
    \]
    where $c_i$, $o_i$, $s_i$, and $r_i$ denote color, opacity, scale, and rotation, respectively, $\mu_i(t)$ is the Gaussian center at time $t \in [0, T\!-\!1]$, and $f_i \in \mathbb{R}^{C}$ is a learnable identity feature~\cite{lin2024gaussian}.
    
    \subsubsection{4DGS Rasterization}
    Following Feature3DGS~\cite{zhou2024feature}, we jointly rasterize photometric and identity feature maps for view $v$ using the differentiable Gaussian rasterizer:
    \[
    (I_v, F_v) = \mathcal{R}(G_t; (R,p)),
    \]
    where $I_v \in \mathbb{R}^{3 \times H \times W}$ is the rendered RGB image, $F_v \in \mathbb{R}^{C \times H \times W}$ is the rendered identity feature map, and $(R,p)$ denotes the camera pose, H, W being image height and width, respectively.
    We decode the rendered feature map into per-pixel instance logits using an MLP decoder~\cite{lin2024gaussian}:
    \[
    L_v = \mathrm{MLP}(F_v) \in \mathbb{R}^{K \times H \times W},
    \]
    where $K$ is the maximum number of object instances.  
    The per-pixel instance probability map is then obtained via a softmax operation:
    \[
    P_v = \mathrm{Softmax}(L_v).
    \]
    
    \begin{figure*}[!t]
        \vspace{0mm}
        \centering
        \includegraphics[width=1.0\linewidth]{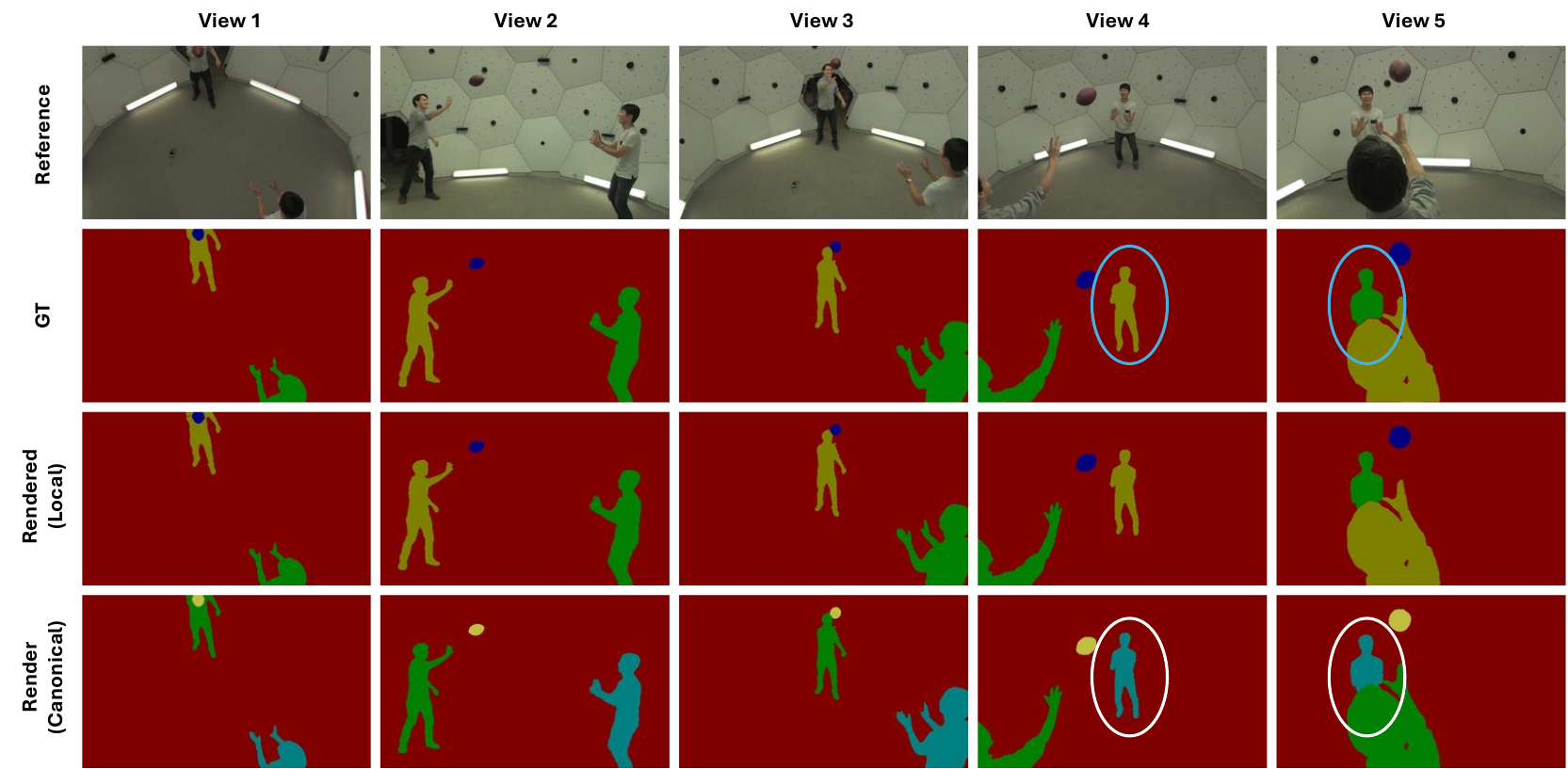}
        \caption{ 
        Example of \textbf{local} and \textbf{canonical} instance maps produced by our method. Because each video stream is segmented independently, SAM3 annotations contain conflicting labels across views (blue circle). Our method resolves this by learning cross-view label permutations to produce consistent canonical labels (white circles). Our differentiable permutation module maps canonical labels ($4^{th}$ row) to view-specific local labels ($3^{rd}$ row), enabling direct supervision despite cross-view inconsistencies. }
         \vspace*{0mm}
        \label{fig:label_permutation}
    \end{figure*}

    \subsubsection{Preprocessing}
    
    Our method takes as input multi-view RGB videos, per-video segmentation maps, and camera poses. 
    To obtain per-video segmentation maps, we apply a video segmentation model (SAM3)~\cite{carion2025sam3} independently to each video stream. Note that the predicted instance labels are not consistent across views (Fig.~\ref{fig:label_permutation}), since the segmentation model processes each video separately.
    
    \subsubsection{Per-Video Latent Permutation}
    To resolve cross-view label inconsistencies, we introduce an implicit learnable label-matching module. 
    Our approach is inspired by SuperGlue~\cite{Sarlin2020superglue}, which solves feature correspondence problems using permutation matrices obtained via differentiable Sinkhorn normalization.
    For each video $v$, we introduce a learnable permutation latent $Z_v \in \mathbb{R}^{K \times K}$
    which represents the correspondence between view-specific (\textit{local}) labels and shared \textit{canonical} labels. 
    Since an unconstrained latent permutation parameters may not represent a valid correspondence matrix, we apply a differentiable Sinkhorn layer to enforce a soft one-to-one assignment:
    
    \begin{align}
    S_v^{(0)} &= \exp(Z_v); \\ 
    S_v^{(i)} &= \mathcal{T}_c\big(\mathcal{T}_r(S_v^{(i-1)})\big),
    \end{align}
    where $\mathcal{T}_r$ and $\mathcal{T}_c$ denote row and column normalization operations. 
    The resulting matrix $S_v$ approximates a permutation matrix.
    The per-video permutation matrix is used to map canonical labels to view-specific labels:
    \[
    L_v^{\text{local}} = S_v \, L^{\text{canon}} .
    \]
    Because the permutation is produced through Sinkhorn normalization, this label remapping process remains fully differentiable while requiring only a small number of learnable parameters ($K \times K$ per video).

    \subsubsection{Progressive Learning}
    
    Jointly optimizing canonical identity features $f_i$ and per-video permutation latents $Z_v$ from scratch is ambiguous, since the canonical label space is initially ungrounded. To stabilize training, we adopt a progressive learning strategy.
    
    We first optimize the Gaussian identity feature fields $f_i$ using only a single reference view. For other views, we detach gradients to $f_i$ while still optimizing their permutation latents. We progressively activate these views for identity field updates once their permutation correspondences are verified with a sufficiently high confidence score. Specifically, we maintain an active view set $V_{\text{active}}$ for updating both $f_i$ and $Z_v$, initialized as $V_{\text{active}}=\{v_{\text{ref}}\}$ and updated by adding any view $v \in V_{\text{all}}$ whose confidence exceeds a threshold:
    \[
    V_{\text{active}} \leftarrow V_{\text{active}} \cup \{\, v \in V_{\text{all}} \mid S(v) > S_{\text{th}} \,\}.
    \]
    Here $S(v)$ is a permutation confidence score computed as the pixel overlap between projected canonical labels and local segmentation labels after applying the current permutation. This progressive activation only affects the segmentation supervision; all views are always used for photometric reconstruction.
    
    \subsubsection{Unseen Object Masking}
    Although SAM3 video segmentation is generally reliable, it can miss fast-moving objects (e.g., the white-circled region in Fig.~\ref{fig:instance_rendering}). In such cases, the rasterized identity map may contain canonical labels that are absent in the corresponding per-view segmentation. We treat pixels whose rasterized labels cannot be matched to any local labels as \emph{unseen} and mask them out when supervising the identity feature fields. This prevents spurious gradients and reduces label ambiguity in regions where objects are missing from the segmentation.
    
    \subsubsection{Loss Functions}
    We jointly optimize segmentation supervision and photometric reconstruction:
    \[
    \mathcal{L} =
    \mathcal{L}_{CE}
    +
    \lambda_1 \mathcal{L}_{L1}
    +
    \lambda_2 \mathcal{L}_{SSIM},
    \]
    where $\mathcal{L}_{CE}$ is the cross-entropy loss on the predicted instance probability maps, and $\mathcal{L}_{L1}$ and $\mathcal{L}_{SSIM}$ are standard photometric rendering losses.

    \section{Optimization with Instance-Decomposed Motion Bases}
    
    \subsection{Instance-Decomposed Motion Bases}
    
    \subsubsection{Per-Instance Motion Bases}
    
    Directly optimizing trajectories for all individual 4D Gaussians is highly ill-posed due to the dynamic nature of the scene. As a result, optimization-based multi-video 4DGS methods are notoriously slow~\cite{som2024, luiten2023dynamic3dgs}, often requiring several hours of training. We leverage motion scaffold structure introduced by MoSca~\cite{lei2025mosca} to model per-instance motion bases for blending the motion of nearby 4D Gaussians. These motion bases improve both the stability and efficiency of 4DGS optimization \cite{lei2025mosca}.

    Given the instance-embedded 4DGS representation, we group Gaussians according to their instance labels and initialize motion bases for each dynamic object instance. First, we randomly sample motion bases from the initialized set of Gaussians. Since we perform sequential optimization starting from the initial time step, the trajectory of each motion base $B_i = (R_i, t_i)$ is initially defined only at $t=0$.
    
    \subsubsection{Motion Blending}
    Each 4D Gaussian $G_i$ is attached to nearby motion bases that share the same instance label. Specifically,
    \[
    \mathcal{B}_i = \{ B_j \mid B_j \in \text{KNN}(G_i), \ \text{Label}(B_j)=\text{Label}(G_i) \}.
    \]
    During optimization, we sequentially update the trajectories of the motion bases at the next time step. Instead of directly optimizing trajectories for all Gaussians, we only optimize the motion bases and propagate their motion to attached Gaussians through motion blending. The trajectory of each Gaussian is obtained by blending the motions of its attached bases. $T_i = \text{DQB}(B_i, w_{ij})$, where $\text{DQB}(\cdot)$ denotes dual-quaternion blending \cite{lei2025mosca} of $SE(3)$ poses and $w_{ij}$ are the blending weights.
    \subsection{Optimization}
    
    \subsubsection{Instance-Aware Rigidity Loss}
    
    We define a local rigidity constraint for each instance to enforce coherent motion among nearby Gaussians. Following Dynamic3DGS and Shape-of-Motion~\cite{luiten2023dynamic3dgs, som2024}, the rigidity loss consists of coordinate, length, and isometry terms:
    
    \[
    \mathcal{L}_\text{rigidity}
    =
    \mathcal{L}_\text{coord}
    +
    \mathcal{L}_\text{len}
    +
    \mathcal{L}_\text{iso}.
    \]
    
    \subsubsection{Optimization Pipeline}
    
    We jointly optimize motion bases together with all Gaussian attributes during training. The instance-aware rigidity loss is applied alongside photometric rendering losses. Thanks to instance-decomposed motion bases, which provide a low-parameter motion model, our method enables efficient optimization.
    
    \[
    \mathcal{L}
    =
    \mathcal{L}_\text{coord}
    +
    \mathcal{L}_\text{len}
    +
    \mathcal{L}_\text{iso}
    +
    \mathcal{L}_\text{L1}
    +
    \mathcal{L}_\text{SSIM}.
    \]

\begin{table}[t]
\centering
\small
\setlength{\tabcolsep}{4pt}
\renewcommand{\arraystretch}{1.12}
\caption{Photometric rendering comparison across datasets.
$\checkmark$ indicates support for the corresponding capability, where \textit{Tracking} denotes long-term trajectory tracking and \textit{Instance} denotes instance-decomposed reconstruction. $\triangle$ denotes tracking via implicit deformation field. We also report training time in minutes.}
\label{tab:photometric_multidataset_pan_neural}
\resizebox{\textwidth}{!}{%
\begin{tabular}{c | cc | ccccc | ccccc}
\toprule
& \multicolumn{2}{c|}{Capability}
& \multicolumn{10}{c}{Dataset} \\
\cmidrule(lr){2-3}
\cmidrule(lr){4-13}

Method
& Tracking
& Instance
& \multicolumn{5}{c|}{PanopticStudio}
& \multicolumn{5}{c}{Neural3DV} \\

& &
& PSNR$\uparrow$ & SSIM$\uparrow$ & LPIPS$\downarrow$ & mIoU$\uparrow$ & Time$\downarrow$
& PSNR$\uparrow$ & SSIM$\uparrow$ & LPIPS$\downarrow$ & mIoU$\uparrow$ & Time$\downarrow$ \\

\midrule

SpacetimeGaussians~\cite{li2024stgs}
& $\times$ & $\times$
& 26.99 & \textbf{0.9211} & \textbf{0.0744} & $\times$ & 28.16
& 28.04 & 0.9364 & 0.0569 & $\times$ & 50.04 \\

Dynamic3DGaussians~\cite{luiten2023dynamic3dgs}
& $\checkmark$ & $\times$
& 26.64 & 0.8548 & 0.2832 & $\times$ & 157.09
& 28.26 & 0.9029 & 0.1785 & $\times$ & - \\

SA4D~\cite{ji2024segment4dgaussians}
& $\triangle$ & $\checkmark$
& 21.42 & 0.7674 & 0.3588 & 0.4618 & \textbf{23.36}
& 28.24 & 0.9290 & 0.1454 & 0.7263 & 61.78 \\

TRASE~\cite{li2026trasetrackingfree4dsegmentation}
& $\triangle$ & $\checkmark$
& 26.10 & 0.8990 & 0.1674 & 0.6402 & 102.60
& 29.29 & 0.9278 & 0.1417 & 0.8978 & 244.80 \\

\midrule
Ours
& $\checkmark$ & $\checkmark$
& \textbf{28.36} & 0.9148 & 0.0848 & \textbf{0.9129} & 27.21
& \textbf{30.88} & \textbf{0.9384} & \textbf{0.0492} & \textbf{0.9420} & \textbf{31.03} \\

\bottomrule
\end{tabular}%
}
\end{table}
    
    \begin{figure*}[!t]
        \vspace{0mm}
        \centering
        \includegraphics[width=1.0\linewidth]{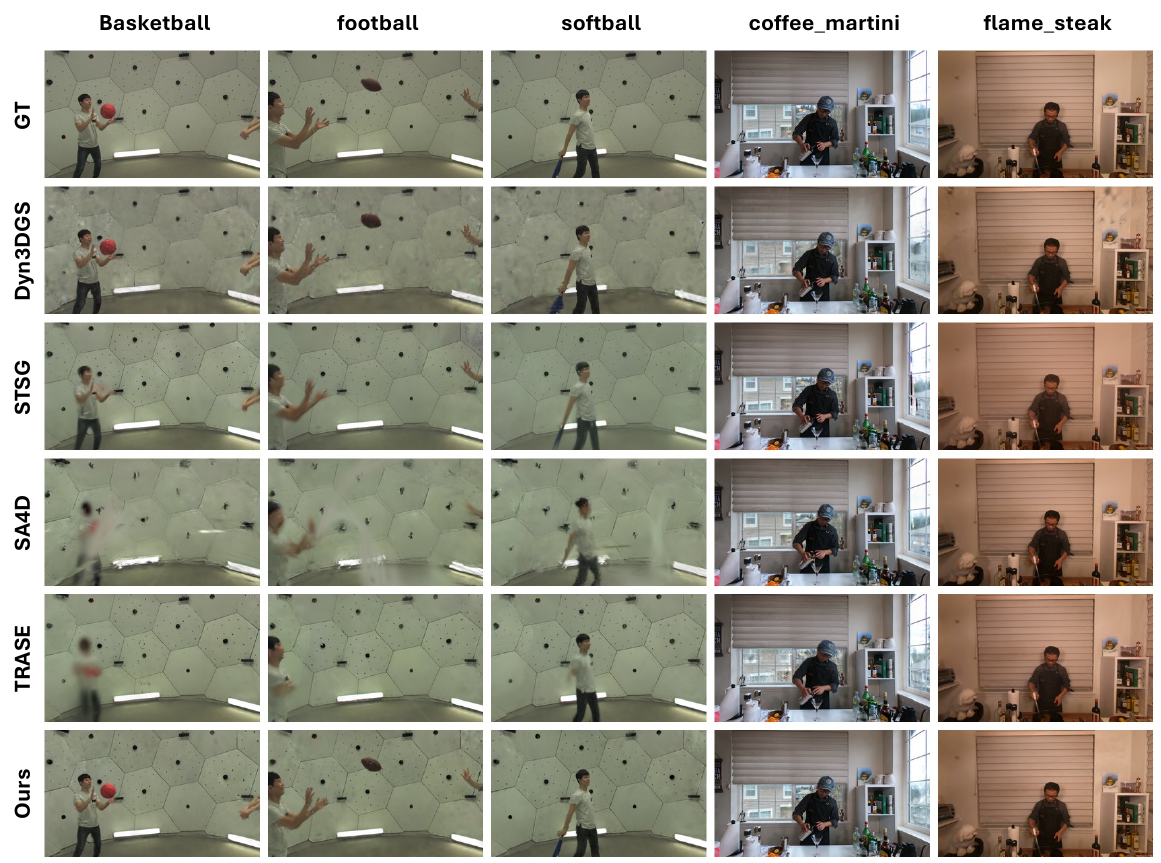}
        \caption{ 
        Qualitative photometric rendering comparison. SA4D and TRASE fail under diverse motions in Panoptic Studio (basketball, football, softball), while our method preserves high-fidelity renderings. We also outperform STSG and Dynamic3DGS, which lack instance-decomposed rendering capability.}
         \vspace*{0mm}
        \label{fig:photometric_rendering}
    \end{figure*}
    
    \begin{figure*}[!t]
        \vspace{0mm}
        \centering
        \includegraphics[width=1.0\linewidth]{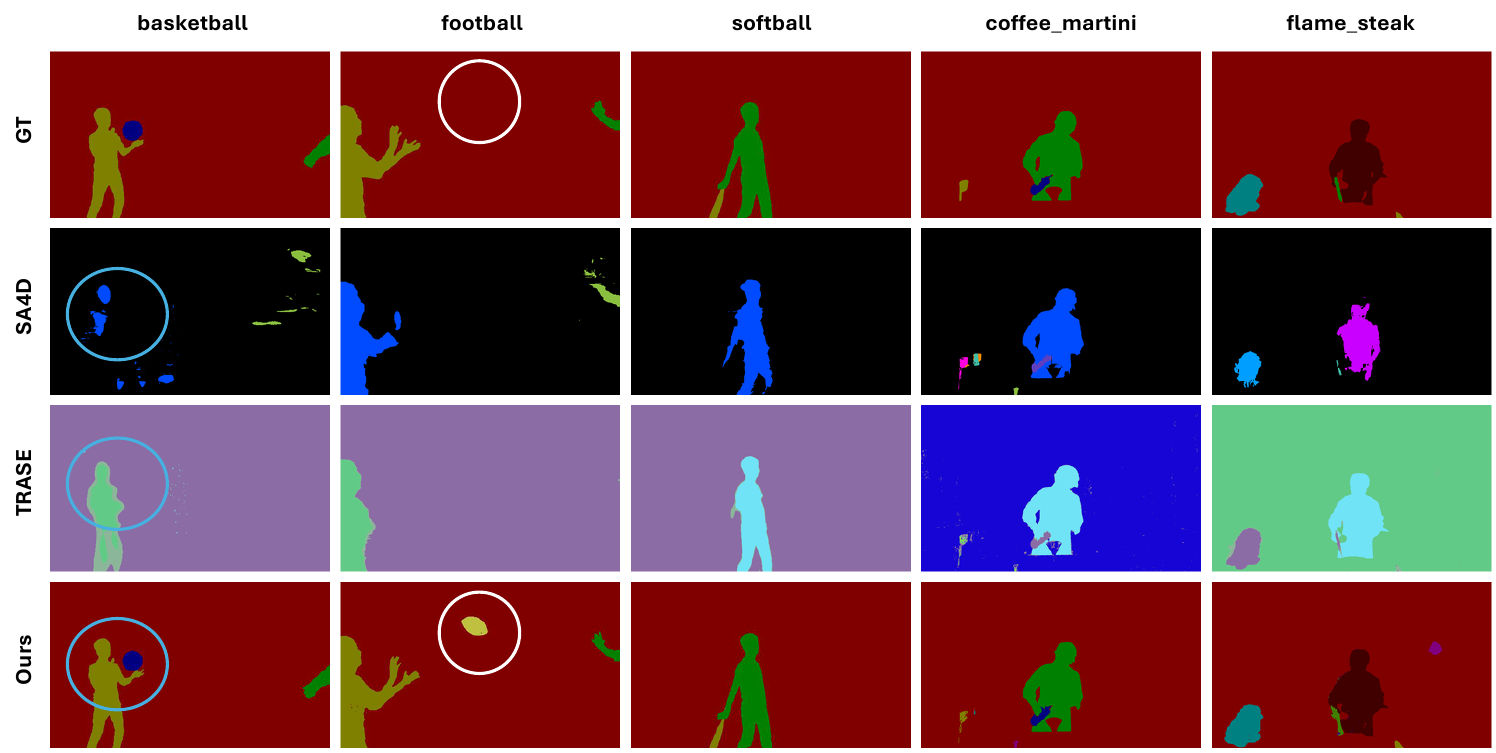}
        \caption{
        Qualitative instance segmentation comparison. SA4D and TRASE are not aligned with test-view labels since they lack cross-view label matching. Our method yields sharper boundaries via direct label supervision, while TRASE shows blurred boundaries and SA4D has incomplete coverage (blue). Although SAM3 may miss the fast-moving ball (white), our unseen-object masking prevents locally untracked objects from degrading our results. }
         \vspace*{0mm}
        \label{fig:instance_rendering}
    \end{figure*}

    \section{Results and Comparisons}
    
    \subsection{Datasets and Tasks}
    
    \subsubsection{Datasets}
    We evaluate our method on \textbf{Panoptic Studio} \cite{joo2015pstudio} and \textbf{Neural3DV} \cite{li2022neural3dv} datasets. The Panoptic Studio dataset contains rich motion of people engaged in multiple activities. We use four scenes from the Panoptic Studio dataset, which include \textit{basketball}, \textit{football}, \textit{tennis}, and \textit{juggle}. Each scene consists of 150 frames at 30 FPS, with 31 synchronized video views. As in Dynamic3DGS~\cite{luiten2023dynamic3dgs}, we use four cameras as test cameras and the other 27 cameras as training cameras. The Neural3DV dataset shows dynamic scenes with a man performing various cooking activities. The Neural3DV dataset consists of five scenes: \textit{coffee\_martini}, \textit{sear\_steak}, \textit{flame\_steak}, \textit{cook\_spinach}, and \textit{cut\_roasted\_beef}. Each scene consists of 300 frames at 30 FPS, with 19--21 synchronized cameras.
    
    To comprehensively evaluate our method, we consider three aspects: (1) photometric rendering, (2) dynamic segmentation, and (3) instance segmentation. We generate semantic annotations for these datasets using SAM3 \cite{carion2025sam3}. Note that each video is processed independently by SAM3, resulting in label spaces that are not consistent across views.
    
    \subsection{Photometric Rendering}
    
    \subsubsection{Evaluation Protocol and Baselines}
    For photometric rendering, we use the standard rendering metrics PSNR, SSIM~\cite{wang2004ssim}, and LPIPS~\cite{zhang2018lpips} to evaluate rendering quality. We compare our algorithm to Dynamic3DGS~\cite{luiten2023dynamic3dgs}, Spacetime Gaussians~\cite{li2024stgs}, SegmentAny4DGaussians~\cite{ji2024segment4dgaussians}, and TRASE~\cite{li2026trasetrackingfree4dsegmentation}. Each of these methods represents a different paradigm of 4D Gaussian splatting. Dynamic3DGS optimizes trajectories of long-term persistent Gaussians with sequential processing of multi-view videos. Spacetime Gaussians optimizes 4D Gaussian primitives defined over a short temporal duration around a reference time. SegmentAny4DGaussians and TRASE use deformable 3DGS to model deformations of reference 3D Gaussians across all timesteps.
    
    \subsubsection{Results}
    We show qualitative and quantitative comparisons in Figure~\ref{fig:photometric_rendering} and Table~\ref{tab:photometric_multidataset_pan_neural}. As shown in Table~\ref{tab:photometric_multidataset_pan_neural}, our method achieves the best rendering performance on both the Neural3DV and Panoptic Studio datasets, which is also evident in the visual results of Figure~\ref{fig:photometric_rendering}. Spacetime Gaussians fails to reconstruct fast-moving objects such as balls while reconstructing people with reasonable quality. We attribute this limitation to the short-term 4D primitives used in STGS, which cannot consistently model long-term trajectories of fast-moving objects. SA4D and TRASE show blurred reconstruction quality because they model motion using deformation-based 3DGS. While their MLP-based deformation modeling performs well for small deformations, as in Neural3DV, it has limited capability to capture long-term trajectories. Dynamic3DGS produces visually reasonable results on Panoptic Studio scenes but requires significantly longer training time because it optimizes the motion of all Gaussians at every timestep. In contrast, our method employs an instance-level motion scaffold that reduces the number of optimization steps while maintaining stronger trajectory reconstruction capability. 
    

    \subsection{Instance Segmentation}

    \subsubsection{Evaluation Protocol and Baselines}
    For instance segmentation, we report mIoU and mAcc, where mIoU is instance-weighted by pixel count. We compare against SegmentAny4DGaussians (SA4D)~\cite{ji2024segment4dgaussians} and TRASE~\cite{li2026trasetrackingfree4dsegmentation}. Following TRASE, we train SA4D using the view closest to the test view, since it cannot directly learn from conflicting multi-view instance labels. TRASE learns instance features via contrastive learning and groups Gaussians using DBSCAN~\cite{ester1996dbscan}. Because per-view segmentation is computed independently, baseline label spaces may not match the test view; thus, we apply test-time label alignment by assigning correspondences that maximize pixel overlap on the test view.

    \subsubsection{Results}
    We present qualitative and quantitative instance segmentation results in Table~\ref{tab:instance_grouped_by_dataset} and Figure~\ref{fig:instance_rendering}. Our method consistently achieves better performance in both mIoU and mAcc on Neural3DV and Panoptic Studio. Overall, these results highlight the advantage of direct supervision over contrastive learning. By explicitly learning cross-view label correspondences through permutation latent learning, our method preserves consistent identities and produces sharper instance boundaries, enabling instance-decomposed 4D Gaussian tracks with long-term identity consistency.

    \subsection{Dynamic Segmentation}
    
    \subsubsection{Evaluation Protocol and Baselines}
    For dynamic segmentation, we use mean Intersection-over-Union (mIoU) and mean pixel accuracy (mAcc), following the evaluation protocol in MaskBenchmark \cite{li2026trasetrackingfree4dsegmentation}. We compare our method with the baselines reported in MaskBenchmark, including Gaussian Grouping \cite{ye2024gaussiangrouping}, SAGA \cite{cen2025segmentany3dgaussians}, DGD \cite{labe2024dgd}, SegmentAny4DGaussians \cite{ji2024segment4dgaussians}, CGC\cite{silva2024cgc}, OpenGaussians \cite{wu2024opengaussian}, and TRASE \cite{li2026trasetrackingfree4dsegmentation}. All methods are evaluated on the foreground dynamic objects in the Neural3DV dataset~\cite{li2022neural3dv}.
    
    \subsubsection{Results}
    We report quantitative dynamic segmentation results in Table~\ref{tab:dynamic_neural3dv_all}. Our method achieves the best overall performance across all scenes in the Neural3DV dataset. SA4D shows limited performance because it cannot leverage multi-view instance labels effectively. Our method slightly outperforms TRASE across the dataset. The lower performance of TRASE can be attributed to the lack of explicit clustering constraints. Because TRASE relies on contrastive feature learning, it learns clustering boundaries implicitly in the feature space. In contrast, our method directly leverages multi-view instance labels, enabling clearer and sharper decision boundaries in both the 2D views and the reconstructed 4D scene representation.

    \begin{table}[t]
    \centering
    \scriptsize
    \setlength{\tabcolsep}{2.5pt}
    \renewcommand{\arraystretch}{1.02}
    \caption{Dynamic segmentation performance on Neural3DV scenes.
    We report Dynamic mIoU and mAcc. Our method shows the best segmentation performance.}
    \label{tab:dynamic_neural3dv_all}
    \resizebox{\columnwidth}{!}{
    \begin{tabular}{p{2.4cm} cc cc cc cc cc}
    \toprule
    Method 
    & \multicolumn{2}{c}{coffee\_martini}
    & \multicolumn{2}{c}{cook\_spinach}
    & \multicolumn{2}{c}{cut\_roasted\_beef}
    & \multicolumn{2}{c}{flame\_steak}
    & \multicolumn{2}{c}{sear\_steak} \\
    \cmidrule(lr){2-3}
    \cmidrule(lr){4-5}
    \cmidrule(lr){6-7}
    \cmidrule(lr){8-9}
    \cmidrule(lr){10-11}
    & mIoU$\uparrow$ & mAcc$\uparrow$
    & mIoU$\uparrow$ & mAcc$\uparrow$
    & mIoU$\uparrow$ & mAcc$\uparrow$
    & mIoU$\uparrow$ & mAcc$\uparrow$
    & mIoU$\uparrow$ & mAcc$\uparrow$ \\
    \midrule
    
    GaussianGrouping
    & 0.8295 & 0.9890
    & 0.8974 & 0.9941
    & 0.9512 & 0.9972
    & 0.8279 & 0.9900
    & 0.9261 & 0.9959 \\
    
    SAGA
    & 0.2201 & 0.8081
    & 0.8125 & 0.9881
    & 0.6982 & 0.9767
    & 0.8439 & 0.9911
    & 0.8959 & 0.9941 \\
    
    SA4D
    & 0.8583 & 0.9910
    & 0.8987 & 0.9941
    & 0.8645 & 0.9914
    & 0.8898 & 0.9940
    & 0.9047 & 0.9948 \\
    
    DGD
    & 0.7875 & 0.9865
    & 0.8150 & 0.9883
    & 0.8170 & 0.9877
    & 0.6771 & 0.9776
    & 0.7638 & 0.9854 \\
    
    CGC 
    & 0.8359 & 0.9895
    & 0.9034 & 0.9945
    & 0.9052 & 0.9943
    & 0.8530 & 0.9920
    & 0.8794 & 0.9934 \\
    
    OpenGaussian
    & 0.8254 & 0.9896
    & 0.6336 & 0.9798
    & 0.9115 & 0.9951
    & 0.8199 & 0.9907
    & 0.8986 & 0.9943 \\
    
    TRASE
    & 0.9120 & 0.9948
    & 0.9129 & 0.9951
    & 0.9103 & 0.9947
    & 0.8716 & 0.9930
    & 0.9044 & 0.9947 \\
    
    \midrule
    
    Ours
    & \textbf{0.9667} & \textbf{0.9980}
    & \textbf{0.9420} & \textbf{0.9954}
    & \textbf{0.9760} & \textbf{0.9964}
    & \textbf{0.9591} & \textbf{0.9909}
    & \textbf{0.9519} & \textbf{0.9968} \\
    
    \bottomrule
    \end{tabular}
    }
    \end{table}

    \begin{table*}[t]
    \centering
    \small
    \setlength{\tabcolsep}{3.2pt}
    \renewcommand{\arraystretch}{1.05}
    \caption{Instance segmentation performance across scenes.
We report instance mIoU and mAcc. While TRASE performs competitively on Neural3DV scenes (\textit{flame\_steak}, \textit{coffee\_martini}), its performance drops on Panoptic Studio scenes. In contrast, our method consistently achieves high-fidelity instance segmentation across both datasets.}
    \label{tab:instance_grouped_by_dataset}
    \resizebox{\linewidth}{!}{%
    \begin{tabular}{@{}l cc cc cc cc cc cc@{}}
    \toprule
    Method
    & \multicolumn{2}{c}{basketball}
    & \multicolumn{2}{c}{football}
    & \multicolumn{2}{c}{tennis}
    & \multicolumn{2}{c}{softball}
    & \multicolumn{2}{c}{coffee\_martini}
    & \multicolumn{2}{c}{flame\_steak} \\
    \cmidrule(lr){2-3}\cmidrule(lr){4-5}
    \cmidrule(lr){6-7}\cmidrule(lr){8-9}
    \cmidrule(lr){10-11}\cmidrule(lr){12-13}
    & mIoU$\uparrow$ & mAcc$\uparrow$
    & mIoU$\uparrow$ & mAcc$\uparrow$
    & mIoU$\uparrow$ & mAcc$\uparrow$
    & mIoU$\uparrow$ & mAcc$\uparrow$
    & mIoU$\uparrow$ & mAcc$\uparrow$
    & mIoU$\uparrow$ & mAcc$\uparrow$ \\
    \midrule
    
    SA4D 
    & 0.3789 & 0.9418
    & 0.5029 & 0.9536
    & 0.4892 & 0.9693
    & 0.4762 & 0.9721
    & 0.9392 & 0.9966
    & 0.5134 & 0.9640 \\

    TRASE
    & 0.5716 & 0.9577
    & 0.6994 & 0.9804
    & 0.6317 & 0.9766
    & 0.6581 & 0.9813
    & 0.9221 & 0.9931
    & 0.8734 & 0.9920 \\
    
    \midrule
    Ours
    & \textbf{0.9314} & \textbf{0.9936}
    & \textbf{0.9269} & \textbf{0.9928}
    & \textbf{0.8769} & \textbf{0.9841}
    & \textbf{0.9165} & \textbf{0.9919}
    & \textbf{0.9851} & \textbf{0.9979}
    & \textbf{0.8988} & \textbf{0.9931} \\
    
    \bottomrule
    \end{tabular}%
    }
    \end{table*}

    \begin{table}[t]
    \centering
    \small
    \setlength{\tabcolsep}{5pt}
    \renewcommand{\arraystretch}{1.12}
    \caption{{\bf Ablation study}: Our instance-decomposed motion bases and Sinkhorn-based label matching jointly contribute to improved rendering quality and instance segmentation performance.}
    \label{tab:ablation_unified}
    \resizebox{\columnwidth}{!}{%
    \begin{tabular}{l ccc cc cc}
    \toprule
    \textbf{Setting}
    & \multicolumn{3}{c}{\textbf{Rendering}}
    & \multicolumn{2}{c}{\textbf{Dynamic Seg.}}
    & \multicolumn{2}{c}{\textbf{Instance Seg.}} \\
    \cmidrule(lr){2-4}\cmidrule(lr){5-6}\cmidrule(lr){7-8}
    & PSNR$\uparrow$ & SSIM$\uparrow$ & LPIPS$\downarrow$
    & mIoU$_{\text{dyn}}\uparrow$ & mAcc$_{\text{dyn}}\uparrow$
    & mIoU$\uparrow$ & mAcc$\uparrow$ \\
    \midrule
    
    \multicolumn{8}{l}{\textbf{Motion Model Ablation}} \\
    \midrule

    w/o Motion Bases
    & 19.2514 & 0.7778 & 0.5435
    & 0.3299 & 0.9303
    & 0.1057 & 0.9080 \\
    
    w/o Instance Grouping
    & 28.1538 & 0.9170 & 0.0858
    & 0.9253 & 0.9926
    & 0.8802 & 0.9882 \\
    
    \midrule
    \multicolumn{8}{l}{\textbf{Instance Model Ablation}} \\
    \midrule
    
    w/o Permutation (No Sinkhorn + No ST + No Track Masking)
    & 24.4539 & 0.8578 & 0.2720
    & 0.0989 & 0.1455
    & 0.1076 & 0.502 \\
    
    w/o Sinkhorn
    & 26.5910 & 0.9108 & 0.0944
    & 0.9107 & 0.9911
    & 0.7050 & 0.9777 \\

    w/o Track Masking
    & 26.8404 & 0.9090 & 0.0975
    & 0.9168 & 0.9919
    & 0.9085 & 0.9914 \\
    
    w/o Progressive Activation
    & 28.4362 & 0.9179 & 0.0851
    & 0.9310 & 0.9927
    & 0.8050 & 0.9787 \\
    
    \midrule
    \textbf{Full Model (Ours)}
    & \textbf{28.5969} & \textbf{0.9163} & \textbf{0.0841}
    & \textbf{0.9280} & \textbf{0.9926}
    & \textbf{0.9190} & \textbf{0.9920} \\
    \bottomrule
    \end{tabular}%
    }
    \vspace*{-0.1in}
    \end{table}
    
    \subsection{Ablation Studies}
    Table~\ref{tab:ablation_unified} presents ablation studies on two key components of our pipeline: motion modeling and instance representation learning. Overall, the results show that instance-decomposed motion bases and permutation latent learning both contribute to improved rendering quality and instance segmentation performance.
    
    \subsubsection{Analysis on Motion Modeling}
We evaluate the effectiveness of instance-decomposed motion bases by ablating motion blending and instance grouping. To test the role of motion bases, we directly optimize trajectories of all 4D Gaussians individually without motion blending. In our full model, each 4DGS motion is blended from bases belonging to the same instance. We also test motion bases without instance grouping to isolate the effect of instance decomposition. As shown in the table, motion bases are crucial for overall 4D reconstruction quality, providing an efficient low-dimensional structure for dynamic motion. The configuration without motion bases resembles Dynamic3DGS~\cite{luiten2023dynamic3dgs}, which optimizes each Gaussian independently and requires many iterations ($\geq 3000$) per timestep, leading to several hours of training. Instance-decomposed motion bases further improve performance by preventing motion mixing across objects, resulting in sharper instance segmentation and improved rendering quality.
    
    \subsubsection{Analysis on Instance Learning}
    We analyze the effectiveness of our per-view permutation learning and progressive training strategy through four ablation configurations. Removing permutation learning severely degrades rendering performance because the model loses the ability to learn a consistent instance field. Without Sinkhorn normalization, the model still learns an instance field using per-video latent variables and an MLP decoder. While this partially resolves label inconsistency (mIoU 0.7050), the performance remains significantly lower than the full model (0.9190), highlighting the importance of Sinkhorn-based permutation learning. Active object masking also improves performance. Removing object masking reduces rendering quality by about 2 PSNR. In addition, removing progressive activation with Hungarian-based verification~\cite{munkres1957assignment} significantly degrades instance mIoU (0.8050 vs.\ 0.9190). This occurs because simultaneously learning instance fields and permutation parameters from all cameras introduces ambiguity during training. Progressive activation mitigates this issue by introducing cameras only after reliable label matching has been established. Combined with per-video segmentation supervision, this strategy enables stable and efficient learning of instance fields from multi-view labels.

    \section{Conclusion and Limitations}
    We present Inst4DGS, an instance-decomposed 4D Gaussian Splatting framework for long-term tracking and reconstruction in dynamic scenes. Inst4DGS resolves cross-view label inconsistencies using per-video permutation latents with differentiable Sinkhorn normalization, enabling direct multi-view supervision with consistent identity preservation. We also introduce an instance-level motion scaffold for efficient long-horizon trajectory optimization. Experiments on Panoptic Studio and Neural3DV show strong performance in photometric rendering, dynamic segmentation, and instance segmentation over concurrent 4DGS baselines. We highlight considerable improvements over prior methods. Inst4DGS has some \textit{limitations}. Since it relies on explicit label supervision, instance masks must be mutually exclusive, which favors sharp boundaries and stable identities but cannot readily model richer semantic hierarchies. Future work includes learning implicit cross-view correspondences and more flexible label spaces.
    
    
    
    %
    %
    \bibliographystyle{splncs04}
    \bibliography{main}

    \end{document}